\documentclass[a4paper]{spie} 

\usepackage{amsmath,amsfonts,amssymb}
\usepackage{graphicx}
\usepackage{subcaption}
\usepackage{hyperref}
\usepackage[margin=2cm]{geometry}
\usepackage{array, booktabs, siunitx}
\hypersetup{colorlinks=true,linkcolor=[rgb]{0.1,0.3,0.9},urlcolor=[rgb]{0.2,0.2,0.2},citecolor=[rgb]{0.1,0.3,0.9}}

\title{Deep Learning-Based Regional White Matter Hyperintensity Mapping as a Robust Biomarker for Alzheimer’s Disease}

\author[]{Julia Machnio}
\author[]{Mads Nielsen}
\author[]{Mostafa Mehdipour Ghazi}
\affil[]{Department of Computer Science, University of Copenhagen, Copenhagen, Denmark}

\begin{document} 
\maketitle

\begin{abstract}
White matter hyperintensities (WMH) are key imaging markers in cognitive aging, Alzheimer’s disease (AD), and related dementias. Although automated methods for WMH segmentation have advanced, most provide only global lesion load and overlook their spatial distribution across distinct white matter regions. We propose a deep learning framework for robust WMH segmentation and localization, evaluated across public datasets and an independent Alzheimer’s Disease Neuroimaging Initiative (ADNI) cohort. Our results show that the predicted lesion loads are in line with the reference WMH estimates, confirming the robustness to variations in lesion load, acquisition, and demographics. Beyond accurate segmentation, we quantify WMH load within anatomically defined regions and combine these measures with brain structure volumes to assess diagnostic value. Regional WMH volumes consistently outperform global lesion burden for disease classification, and integration with brain atrophy metrics further improves performance, reaching area under the curve (AUC) values up to 0.97. Several spatially distinct regions, particularly within anterior white matter tracts, are reproducibly associated with diagnostic status, indicating localized vulnerability in AD. These results highlight the added value of regional WMH quantification. Incorporating localized lesion metrics alongside atrophy markers may enhance early diagnosis and stratification in neurodegenerative disorders.
\end{abstract}


\section{INTRODUCTION}

White matter hyperintensities (WMHs) appear as hypointense on T1-weighted MRI and hyperintense on FLAIR, and are commonly observed in older adults. They are linked to cerebral small vessel disease, aging, and cognitive decline [\citenum{zheng2023analysis}]. WMHs are increasingly recognized as an imaging marker in Alzheimer’s disease (AD) and related neurodegenerative conditions [\citenum{brickman2015reconsidering}]. While the overall WMH load has been extensively studied, recent work suggests that the spatial distribution of WMHs across white matter regions provides additional insight into cognitive status and disease progression [\citenum{dadar2022white}].

Traditional approaches to regional WMH analysis rely on labor-intensive manual segmentations and region-of-interest delineations, which limit their scalability for large studies or clinical applications. These methods are not only time-consuming but also subject to inter-rater variability, making it challenging to apply them across multi-center cohorts. Recent advances in deep learning have made it possible to automate the segmentation of both WMHs and anatomical regions [\citenum{machnio2025deep,machnio2025towards}]. Such models enable fast, reproducible, and standardized extraction of region-specific lesion volumes from MRI, opening the door to large-scale studies of spatial WMH patterns.

\begin{figure}[t]
\vspace{0.5cm}
\centering
\includegraphics[width=0.98\linewidth]{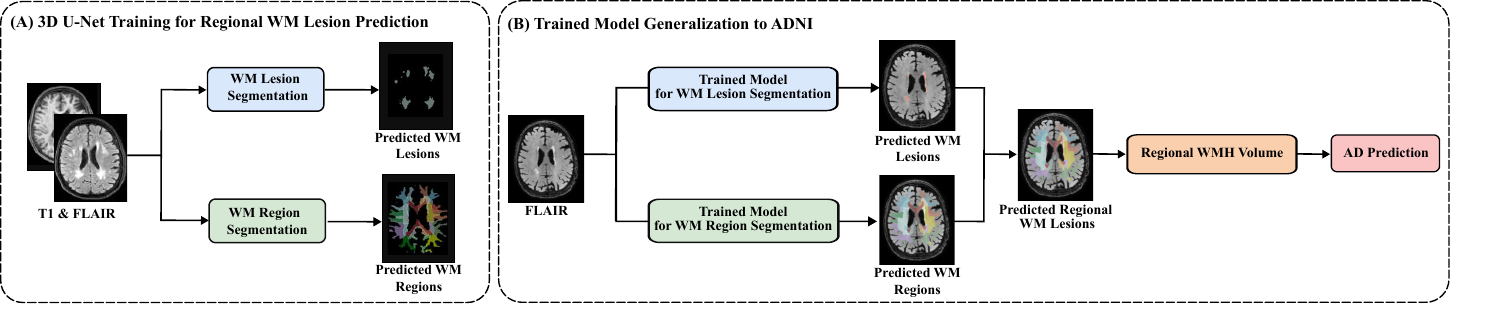}
\caption{Overview of the proposed pipeline for AD status prediction. (A) Regional WMH lesion and white matter region maps are obtained using 3D U-Net models [\citenum{machnio2025deep,machnio2025towards}] trained on the WMH Challenge dataset. (B) The trained models are applied to ADNI scans to extract local and global WMH volumes, which are subsequently used for downstream AD status prediction.}
\label{fig:method}
\end{figure}

In this study, we investigate the predictive value of regional WMH features derived from deep learning-based segmentation pipelines. As shown in Figure~\ref{fig:method}, we train models for WMH and brain region segmentation using the MICCAI 2017 WMH Segmentation Challenge dataset [\citenum{kuijf2019standardized}], and apply them to baseline FLAIR scans from the Alzheimer’s Disease Neuroimaging Initiative (ADNI) [\citenum{weiner2017alzheimer}]. Segmentation performance is assessed by Bland–Altman analysis against standardized WMH volume estimates. We then examine how global versus local WMH features differentiate cognitively normal (CN), mild cognitive impairment (MCI), and AD subjects, and further evaluate whether combining WMH features with brain volumes improves prediction.

Our findings indicate that regional WMH load generalizes well across datasets and enhances classification accuracy compared to global measures alone. Importantly, these results highlight that spatially localized lesions can reveal subtle disease-related alterations not captured by global metrics. This suggests that regional WMH quantification may serve as a complementary imaging biomarker in neurodegenerative research [\citenum{nielsen2025assessing}], with potential to improve both mechanistic understanding and the development of disease prediction tools.

\section{METHODS}

\subsection{Datasets}

We used the MICCAI 2017 WMH Segmentation Challenge dataset [\citenum{kuijf2019standardized}] to train and evaluate our deep learning models for WMH and regional segmentation. This dataset contains co-registered 3D FLAIR and T1-weighted MRI scans from 170 subjects across three cohorts (Utrecht, Amsterdam, and Singapore), with expert WMH annotations provided as reference. We split the dataset into 110 training and 60 testing subjects, and performed five-fold cross-validation for training both lesion and region segmentation models.

Ground-truth labels for white matter region segmentation were derived from the refined JHU MNI White Matter Atlas Type II [\citenum{oishi2009atlas}]. We refined the atlas to delineate 34 anatomically and clinically relevant WM subregions, defined using ontological relationships and neuroanatomical landmarks. To obtain subject-specific region labels, the atlas T1 image was affinely registered to each subject’s native T1 scan, and the resulting transformation was applied to the atlas labels. This yielded anatomically aligned subregion maps in subject space. A detailed description of the preprocessing pipeline is provided in [\citenum{machnio2025deep}].

For external validation and downstream analysis, we used a subset of the ADNI 3 dataset [\citenum{weiner2017alzheimer}]. We selected 552 baseline scans with available 3D FLAIR MRIs, clinical diagnostic labels (CN, MCI, and AD), and standardized WMH volume estimates from the UCD imaging pipeline, combining FLAIR intensity thresholds with anatomical priors [\citenum{decarli2013four}]. All scans had near-isotropic resolution, with an average voxel size of $1 \times 1 \times 1.2$ mm$^3$.

\subsection{WMH Segmentation and Localization}

We used 3D U-Net architectures [\citenum{ronneberger2015u}] for WM lesion and region segmentation [\citenum{machnio2025deep,machnio2025towards}]. To improve model robustness and generalization, we applied MRI-specific data augmentations, including additive and multiplicative noise, bias field distortions, elastic deformations, random rotations, and simulated motion artifacts [\citenum{llambias2024yucca}]. Models were optimized using a composite loss function that combined voxel-wise cross-entropy and Dice–Sørensen losses. 

At inference, predictions were generated independently from FLAIR and T1-weighted scans. The softmax probability maps from both modalities were averaged to fuse outputs, and the final voxel-wise label map was obtained using the \texttt{ArgMax} operation. A full description of preprocessing and training is available in [\citenum{machnio2025towards}].

\subsection{Local WMH Load for AD Status Prediction}

We quantified subject-specific WMH volumes globally and across 34 predefined WM regions. Global WMH volume was defined as the total lesion volume across all regions, whereas local load reflected region-specific lesion loads. For downstream prediction, we used ADNI scans with clinical diagnoses. We formulated three binary classification tasks: AD vs. CN, AD vs. MCI, and MCI vs. CN. 

Four sets of input features were evaluated: (1) global WMH volume, (2) regional WMH volumes, (3) brain structure volumes, including hippocampus, CSF, white matter, and gray matter volumes, and (4) concatenated WMH and volumetric features as multivariate predictors. Linear regression models were trained for classification, and performance was assessed using receiver operating characteristic (ROC) curves and the area under the ROC curve (AUC). This design enabled us to compare the relative predictive value of global versus local WMH features and to evaluate their complementarity with standard brain volumetrics.

\section{RESULTS}

\begin{figure}[t]
\begin{subfigure}{0.31\textwidth}
\centering
\caption*{A}
\vspace{-0.1cm}
\includegraphics[width=\linewidth]{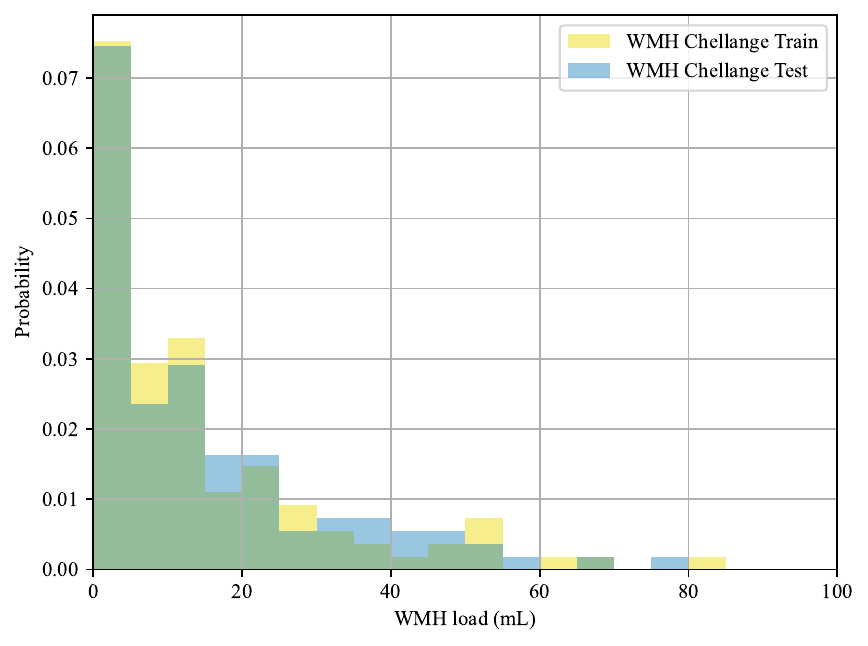}
\end{subfigure}
\hfill
\begin{subfigure}{0.31\textwidth}
\centering
\caption*{B}
\vspace{-0.1cm}
\includegraphics[width=\linewidth]{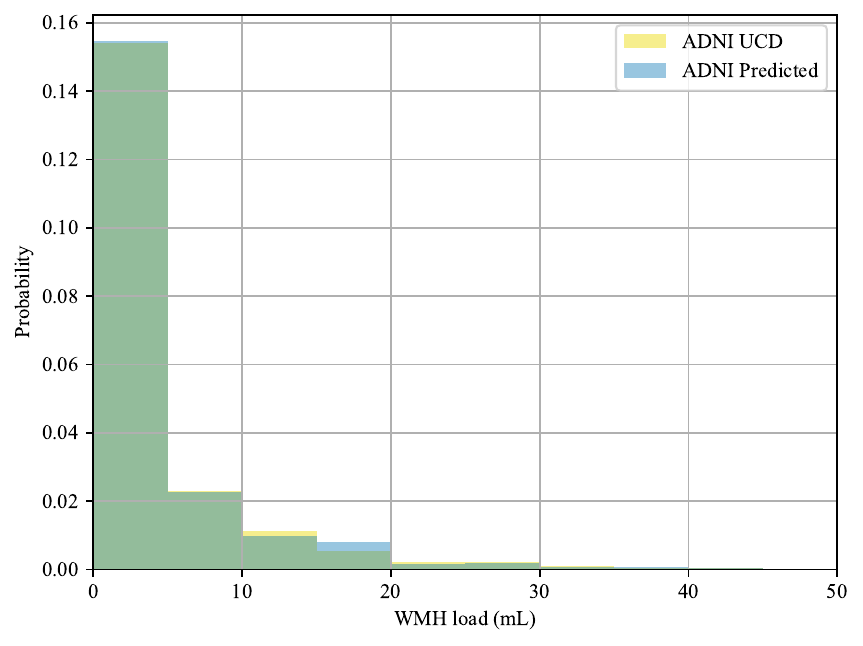}
\end{subfigure}
\hfill
\begin{subfigure}{0.365\textwidth}
\centering
\caption*{C}
\vspace{-0.1cm}
\includegraphics[width=\linewidth]{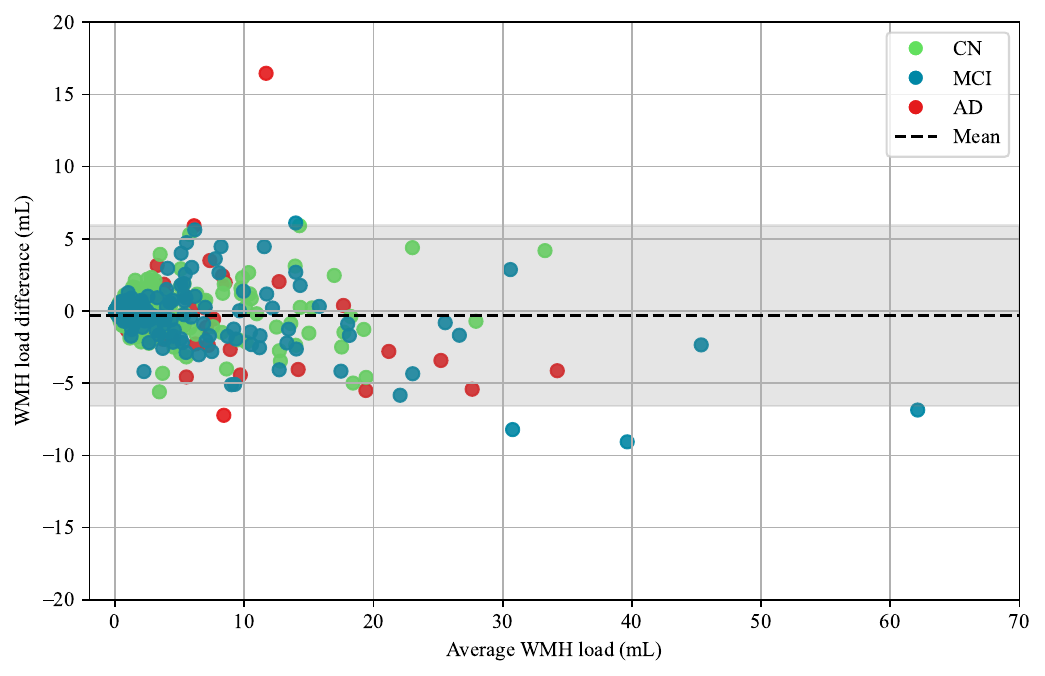}
\end{subfigure}
\caption{(A–B) Distribution of WMH load across datasets: (A) WMH Challenge training and test sets, (B) Our predicted WMH load compared to UCD reference estimates in ADNI. (C) Bland–Altman plot of our predicted versus reference WMH volumes in ADNI, with each point representing a subject colored by diagnostic group.}
\label{fig:wmh_histograms}
\end{figure}

Figure~\ref{fig:wmh_histograms} shows the WMH distributions in the WMH Challenge dataset (A) and the ADNI subset (B). The WMH load in ADNI is substantially lower and less variable, consistent with its recruitment of community-dwelling older adults with relatively mild clinical presentations [\citenum{weiner2017alzheimer}]. In contrast, the WMH Challenge cohorts consisted of hospital-based stroke and memory clinic patients with higher vascular burden and lesion variability [\citenum{kuijf2019standardized}]. Despite this domain shift, predicted WMH load in ADNI closely matched the reference UCD estimates, demonstrating the strong generalization capacity of the segmentation model, even though it was not trained on ADNI data.

\begin{figure}[t]
\vspace{0.5cm}
\centering
\includegraphics[width=\linewidth]{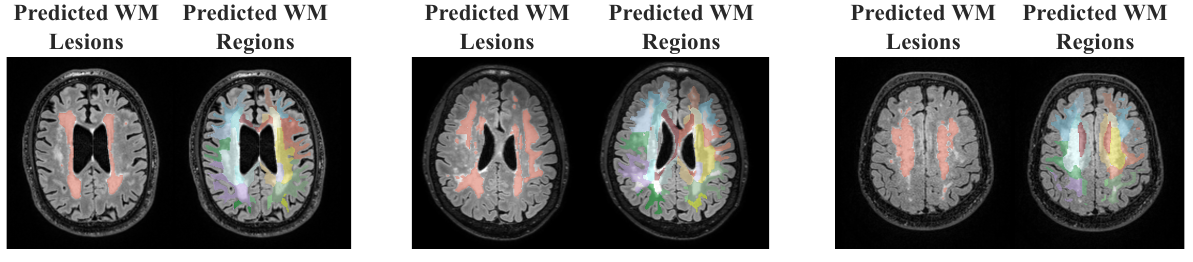}
\caption{Predicted white matter lesions and regional white matter segmentations for three representative ADNI subjects with high WMH loads.}
\label{fig:adni}
\end{figure}

To further quantify this agreement, we performed Bland-Altman analysis of predicted versus reference WMH volumes in ADNI (C). The mean bias was near zero, and most points fell within the 95\% limits of agreement, indicating no systematic over- or under-estimation across the WMH volume range. Model performance was consistent across diagnostic categories (CN, MCI, AD), supporting applicability to diverse clinical populations. Representative examples of lesion and region segmentations for ADNI subjects are shown in Figure~\ref{fig:adni}.

\begin{figure}[t]
\begin{subfigure}{0.33\textwidth}
\centering
\includegraphics[width=\linewidth]{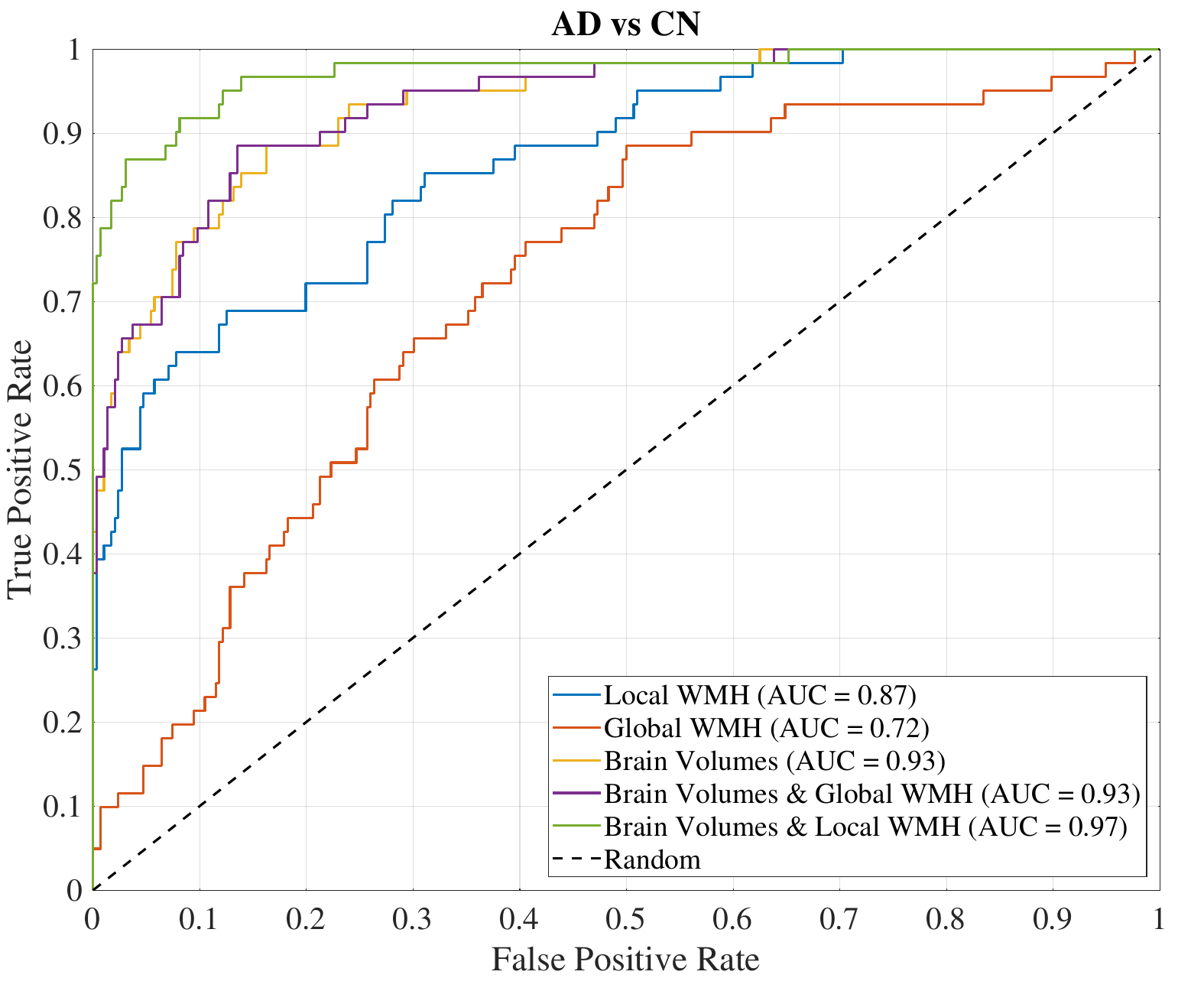}
\end{subfigure}
\begin{subfigure}{0.33\textwidth}
\centering
\includegraphics[width=\linewidth]{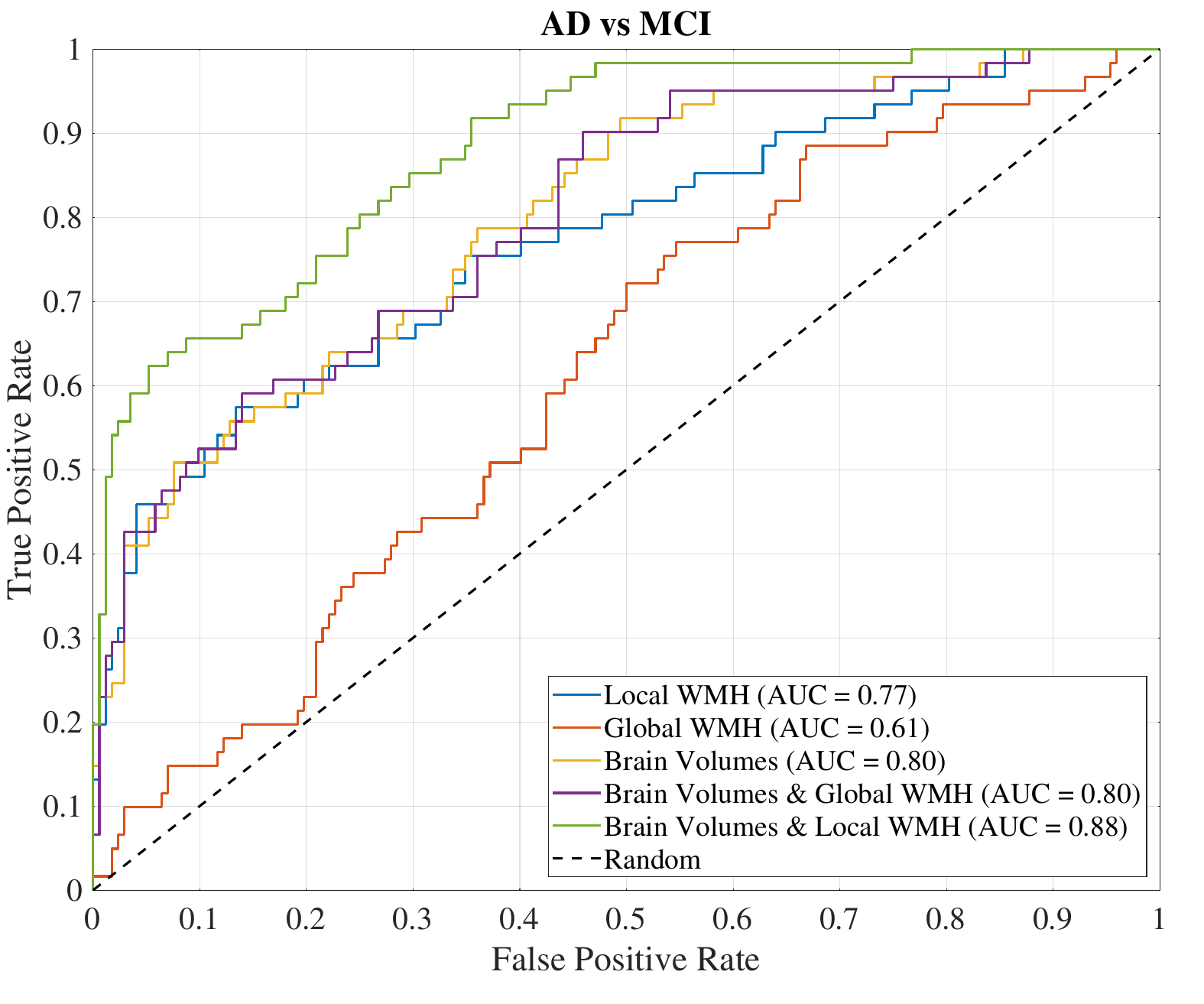}
\end{subfigure}
\begin{subfigure}{0.33\textwidth}
\centering
\includegraphics[width=\linewidth]{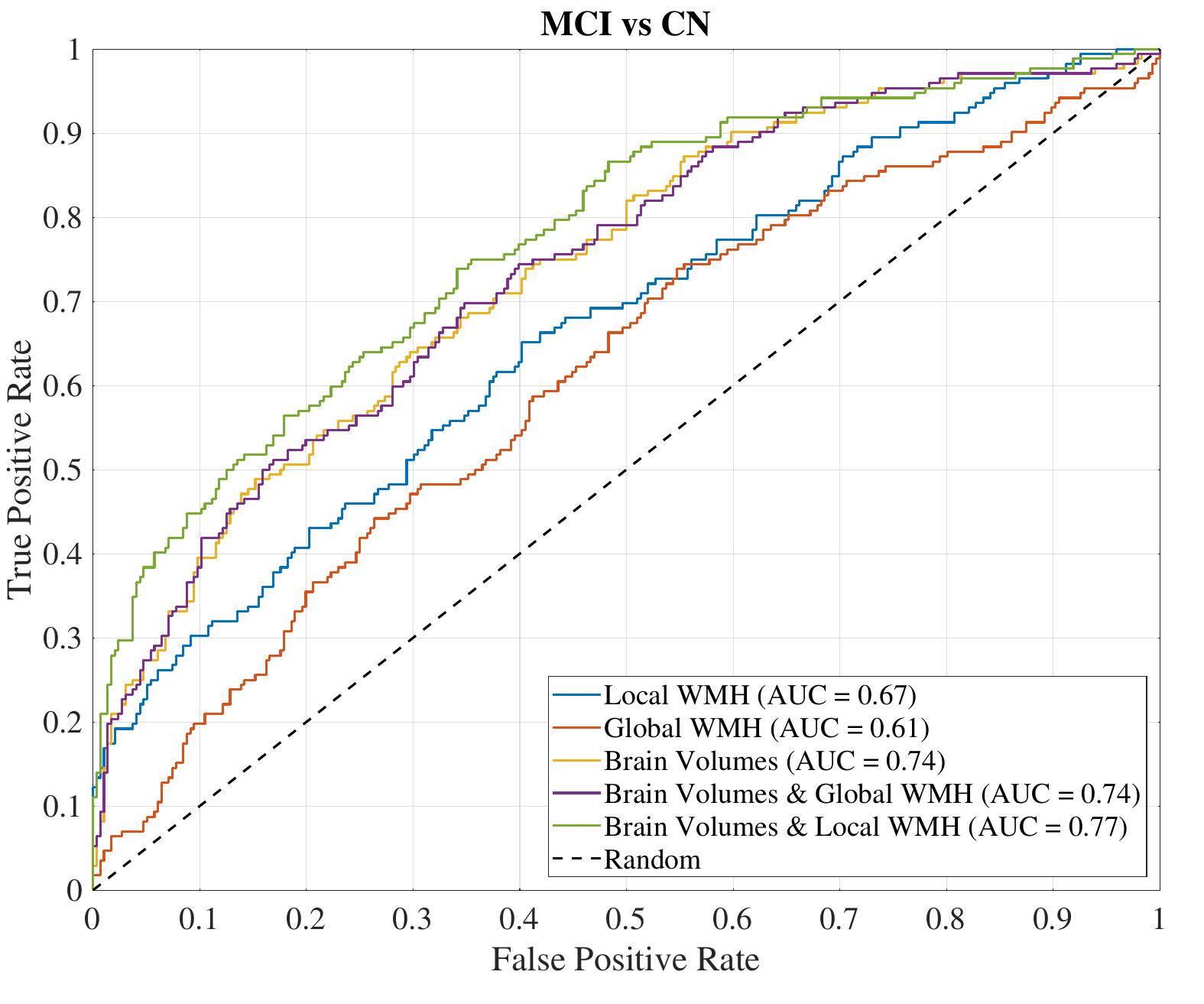}
\end{subfigure}
\caption{ROC curves for AD prediction tasks using different MRI feature sets. Regional WMH consistently outperforms global WMH, and combining local WMH with brain volumetrics further boosts performance.}
\label{fig:roc_curves}
\end{figure}

We next evaluated the predictive value of WMH features for clinical classification. Using three binary tasks, we compared four feature sets: global WMH volume, local WMH volumes, brain structure volumes, and concatenated WMH load and brain volumetry. As shown in Figure~\ref{fig:roc_curves}, regional WMH consistently outperformed global WMH. For example, in the AD vs. CN task, regional WMH load achieved an AUC of 0.87 compared to 0.72 for global WMH load. As shown in the literature [\citenum{mehdipour2024comparative,mehdipour2025fast,nielsen2025assessing}], regional brain volumes are strong predictors of AD and achieve higher accuracy than WMH features. Integrating regional WMH volumes with brain volumetrics further boosted performance, reaching an AUC of 0.97. These improvements were observed across all tasks, highlighting the added predictive value of spatial lesion distribution beyond global burden.

To interpret these results, we examined regression coefficients to identify regions that significantly contributed to prediction ($p < 0.05$). The most discriminative regions were: regions 1 (Superior Parietal Lobule, Angular Gyrus, Supramarginal Gyrus), 7 (Precuneus, Cuneus, Lingual Gyrus), 11 (Fronto-Orbital Cortex), 22 (Inferior Frontal Gyrus), 29 (Superior Longitudinal Fasciculus, Superior Corona Radiata), and 33 (Posterior Limb of Internal Capsule, Cerebral Peduncle) for AD vs. CN; regions 1 and 22 for AD vs. MCI; and regions 1 and 24 (Posterior Cingulate, Precuneus, Lingual Gyrus) for MCI vs. CN.

Several of these regions align with networks known to be vulnerable in AD. Region 1 encompasses parietal areas involved in visuospatial attention, working memory, and multisensory integration functions commonly impaired in early AD [\citenum{wagner2005parietal,jacobs2012parietal}]. Region 22 is implicated in executive function [\citenum{seeley2007dissociable}] and language processing [\citenum{friederici2011brain}]. The occipito-parietal cluster in Region 7 spans the precuneus, a hub for self-referential processing and episodic memory [\citenum{wagner2005parietal,cavanna2006precuneus,buckner2008brain}], as well as the cuneus and lingual gyrus, which are also affected in AD. Regions 29 and 33 cover major association and projection pathways (superior longitudinal fasciculus, corona radiata, internal capsule, cerebral peduncle), reflecting long-range disconnection associated with disease severity [\citenum{bai2009mci,bai2009association,nasrabady2018white}]. Region 24 involves posterior cingulate and precuneus structures, recognized as early sites of neurodegeneration [\citenum{kinkingnehun2008longitudinal,schott2010pcc}].

\section{CONCLUSION}

We introduced a deep learning framework for WMH segmentation and localization and demonstrated its robust generalization to the external ADNI cohort. The predicted lesion loads showed strong concordance with reference WMH volume estimates, highlighting the model's capacity to accommodate domain shifts in lesion load and demographic characteristics.

By deriving region-wise WMH volumes and integrating them with brain structure volumes, we achieved high predictive accuracy across different prediction tasks. Regional WMH consistently outperformed global lesion volume as a marker, and its combination with brain atrophy measures yielded the best results, with AUCs up to 0.97. Several spatially distinct regions, most notably Region 1, were repeatedly linked to diagnostic status, suggesting localized vulnerability of specific white matter pathways in neurodegeneration.

These findings underscore the added value of regional WMH quantification for clinical research and disease stratification. Incorporating such metrics alongside conventional atrophy markers may enhance early detection of Alzheimer’s disease and related disorders, and support more precise characterization of disease mechanisms in future longitudinal studies.

\section*{ACKNOWLEDGMENTS}       

This project is supported by the Pioneer Centre for AI, funded by the Danish National Research Foundation (grant number P1).

\bibliographystyle{spiebib}
\bibliography{references}

\end{document}